\pdfoutput=1

\documentclass[11pt]{article}

\usepackage[review]{ACL2023}

\usepackage{times}
\usepackage{latexsym}
\usepackage[T1]{fontenc}

\usepackage[utf8]{inputenc}

\usepackage{microtype}

\usepackage{inconsolata}
\usepackage{menukeys}

\newcommand\blfootnote[1]{%
  \begingroup
  \renewcommand\thefootnote{}\footnote{#1}%
  \addtocounter{footnote}{-1}%
  \endgroup
}

\title{Enhancing OCR Performance through Post-OCR Models: \\Adopting Glyph Embedding for Improved Correction}

\author{Yung-Hsin Chen$^\ast$ \\ 
    University of Zurich \\
    \texttt{yung-hsin.chen@uzh.ch}
    \And
    Yuli Zhou$^\ast$ \\ 
    University of Zurich \\ 
    \texttt{yuli.zhou@uzh.ch}
}

\begin{document}
\nolinenumbers
\maketitle
\blfootnote{$^\ast$ These authors contributed equally to this work.}
\begin{abstract}
The study investigates the potential of post-OCR models to overcome limitations in OCR models and explores the impact of incorporating glyph embedding on post-OCR correction performance. In this study, we have developed our own post-OCR correction model. The novelty of our approach lies in embedding the OCR output using CharBERT and our unique embedding technique, capturing the visual characteristics of characters. Our findings show that post-OCR correction effectively addresses deficiencies in inferior OCR models, and glyph embedding enables the model to achieve superior results, including the ability to correct individual words.
\end{abstract}
\section{Introduction}
\label{sec:introduction}
\paragraph{}Converting images into digital formats has become increasingly practical for preserving ancient documents~\cite{Avadesh2018CNN-OCR, Narang2020Ancient}, understanding real-time road signs~\cite{Wu2005Detection, Kim2020Application}, multimodal information extraction~\cite{Liu2019Graph, Sun2021RpBERT} and evaluating text-guided image generation~\cite{Petsiuk2022HumanEval, Zhang2023MagicBrush, Rodriguez2023OCR-VQGAN}, just to name a few.
These digital representations can be further processed using language models to extract underlying features. A high-quality OCR model can benefit various domains. However, the performance of optical character recognition (OCR) is often limited due to factors such as image quality or inherent model limitations.\par

This study seeks to explore the potential of post-OCR correction, specifically examining whether it can compensate for the deficiencies of OCR models. The study comprises two main components. Firstly, we evaluated a range of OCR models and their post-OCR correction methods using language models to determine whether the correction process could mitigate weaknesses in the OCR results. Secondly, we introduce our own embedding and post-OCR correction model, leveraging CharBERT \cite{ma-etal-2020-charbert} for character embedding, as well as a custom glyph embedding developed by our team. This glyph embedding captures the visual characteristics of the characters, rather than solely focusing on semantic aspects like most conventional embeddings.

\section{Data}
\paragraph{}In the initial phase of our study, we employ the extensively researched and widely employed ICDAR 2013 dataset as our primary data source for OCR tasks. The results obtained from this phase will subsequently serve as inputs for the next step, where we evaluate the performance of our post-OCR correction model. Additionally, to facilitate training of the glyph embedding in our post-OCR correction model, we rely on the Chars74K dataset\footnote{Available at \url{http://www.ee.surrey.ac.uk/CVSSP/demos/chars74k/}}.
\label{sec:data}
\paragraph{}The ICDAR 2013 dataset \cite{6628859} and the recently introduced ICDAR 2023 dataset \cite{long2023icdar} serve as prominent benchmarks for evaluating Optical Character Recognition (OCR) models. \par 
The ICDAR 2013 dataset encompasses a diverse collection of document images, encompassing both printed and handwritten text captured under a variety of conditions, including different languages, font styles, sizes, and distortions. It serves as a robust benchmark for assessing the accuracy and robustness of OCR systems in a wide array of real-world scenarios. \par
In contrast, the ICDAR 2023 dataset expands upon the achievements of its predecessor by introducing additional challenges that push the boundaries of OCR technology. It incorporates more intricate document layouts, degraded text, low-resolution images, and challenging background clutter, with the goal of evaluating OCR model performance in increasingly demanding scenarios.\par
Both datasets will undergo evaluations using three models: EasyOCR\footnote{\url{https://github.com/JaidedAI/EasyOCR}}, PaddleOCR\footnote{\url{https://github.com/PaddlePaddle/PaddleOCR}}, and TrOCR (Transformer-based OCR) \cite{li2021trocr}. The EasyOCR model will be utilized specifically for single word detection box, while the other two models will provide single line detection box functionality. Consequently, the outputs from EasyOCR will consist solely of single words, whereas PaddleOCR will generate output in the form of sentences.
\paragraph{}The Chars74K dataset is employed for training the glyph embedding. This dataset encompasses scene images of English and Kannada characters, although for this particular experiment, only English alphabets are utilized. In addition to the Chars74K dataset, we capture screenshots of Korean and Hebrew characters to serve as samples for the garbage class in open-set classification.\par
For training the post-OCR correction model, the data consists of the outputs obtained from EasyOCR, PaddleOCR, and TrOCR in the initial phase.
\section{Method}
\label{sec:method}
\paragraph{}In this section, we will outline the methodologies employed to conduct the examinations of state-of-the-art (SOTA) models as well as our post-OCR correction model.
\paragraph{SOTA OCR models}In recent years, OCR models have leveraged advanced machine learning and computer vision techniques to achieve remarkable levels of accuracy and efficiency. EasyOCR has emerged as a popular open-source OCR library. Within EasyOCR, the detection execution employs the CRAFT algorithm \cite{baek2019character}, while the recognition model is based on CRNN \cite{shi2016end}. The CRNN model comprises three key components: feature extraction utilizing ResNet \cite{he2016deep} and VGG, sequence labeling employing LSTM, and decoding through CTC. PaddleOCR, another widely adopted OCR framework, is built on the PaddlePaddle deep learning platform and showcases state-of-the-art performance across various text recognition tasks. PaddleOCR supports multiple languages, scene text recognition, and text detection based on PP-OCRv3 \cite{li2021trocr}. Furthermore, TrOCR stands as an OCR model that capitalizes on transformer-based architectures to achieve exceptional recognition accuracy. By harnessing the power of transformers, TrOCR effectively captures contextual information and delivers impressive results across diverse OCR tasks. 
\paragraph{Post-OCR Correction by Language Models}Post-OCR correction using Language Models (LMs) has emerged as a robust methodology for enhancing the accuracy and quality of Optical Character Recognition (OCR) outputs. LMs, such as transformer-based architectures like BERT \cite{devlin2018bert} or GPT \cite{brown2020language}, have revolutionized the field of natural language processing. Within the realm of post-OCR correction, LMs leverage their contextual understanding and language modeling capabilities to detect and rectify errors present in the recognized text. By analyzing the neighboring words, sentence structure, and semantic context, LMs can effectively identify and rectify OCR mistakes, encompassing misspellings, punctuation errors, and grammatical inconsistencies. This approach has demonstrated its efficacy in significantly enhancing OCR accuracy, particularly in scenarios where OCR models face challenges associated with complex layouts, degraded text, or intricate font styles.

\paragraph{}The post-OCR correction model is composed of two parallel encoders and one decoder. The encoders receive inputs in the form of sentences that have been embedded using CharBERT and the glyph embedding that we developed through training.
\paragraph{CharBERT and Glyph Embedding}CharBERT is a pre-trained language model that takes a string as input and produces its corresponding embedding. The model combines BERT embedding with character-level embedding. It undergoes two pretraining tasks: masked word prediction\footnote{Similar to BERT's pretraining task, CharBERT randomly masks words for prediction.} and noise deduction\footnote{CharBERT introduces random noise, such as random characters, into the corpus and trains the model to denoise the text.}. We chose this embedding for our post-OCR correction model because the denoising pretraining task is expected to aid in the correction process.\par
On the other hand, the glyph embedding is trained using open-set classification with a garbage class\footnote{Data categorized as "garbage" refers to information that does not hold significance within the context of our study. Its function lies in training the model to classify irrelevant characters (specifically, non-English alphabets and non-digits in this instance) into a designated category termed the "garbage" class. In the scope of this research, the training data for the garbage class encompasses Hebrew and Korean characters.} and adapted softmax. ResNet18 and ResNet50 models are utilized, with the last fully-connected layer (fc) replaced and an additional fc layer appended for the purpose of fine-tuning. During inference, this additional fc layer is removed to generate an embedding of size 768 (Table \ref{tab:resnet_arch}). In fine-tuning, we experiment with unfreezing the last two fc layers as well as the last four layers. Unlike charBERT, which captures the semantic information of words, glyph embedding represents the visual glyph of characters.\par
\begin{figure}[h]
\includegraphics[width=8cm]{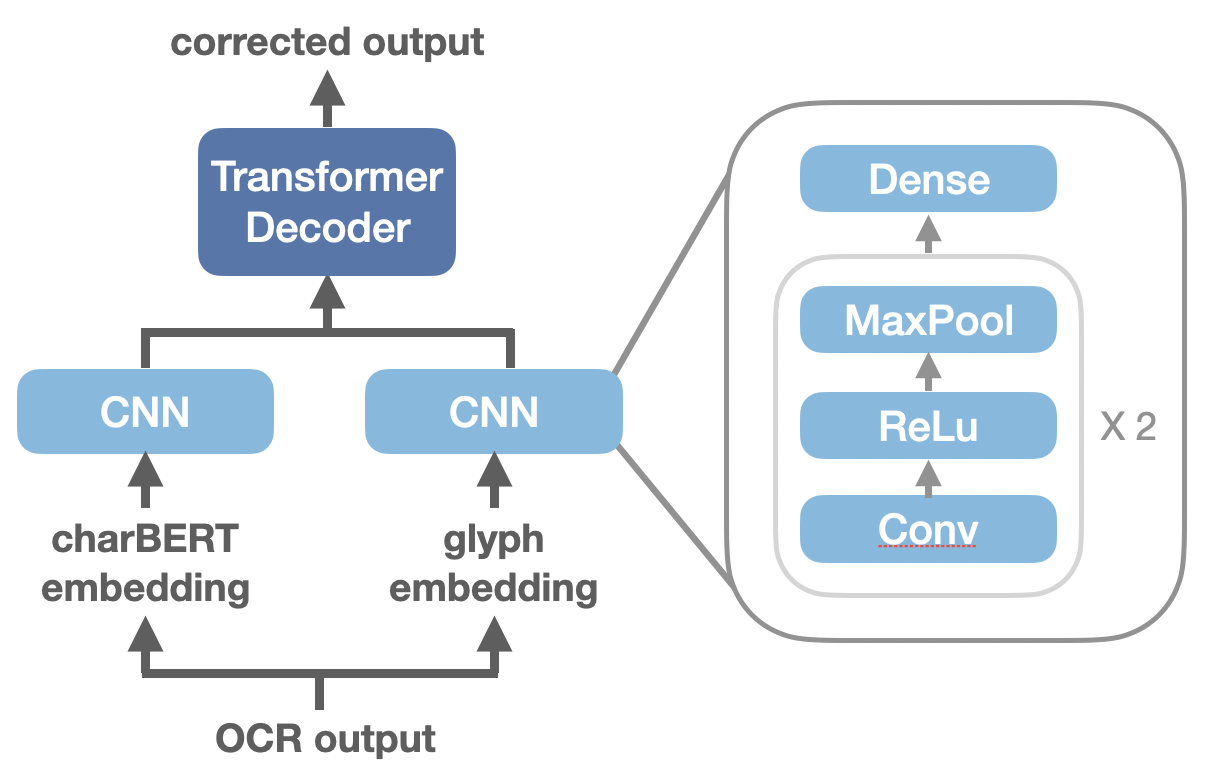}
    \caption{The post-OCR correction model consists of two parallel CNN encoders and a transformer decoder with CharBERT and glyph embedding as inputs.}
    \label{fig:model}
\end{figure}

After fine-tuning, images of the same character are passed through the model to generate embeddings. The averaged embedding is considered as the glyph embedding for that specific character. 
\paragraph{Post-OCR Correction Model}Once the glyph embedding training is complete, it becomes one of the inputs for the post-OCR correction model. This model is composed of two CNN encoders and one transformer decoder (Figure \ref{fig:model}). CNN is a commonly employed architecture for processing character embeddings \cite{kim2016character}, while the transformer decoder offers a significant advantage in handling long-distance dependencies \cite{vaswani2017attention}. Furthermore, the positional encoding within the transformer decoder aids in preserving the spatial information of the input sentence.
\section{Results}
\label{sec:results}
\paragraph{Evaluation Criteria}To assess the recognition performance of the OCR model, we employ a word-level evaluation using full matching. In addition, for line-level evaluation, two metrics are utilized: Character Error Rate (CER) and Word Error Rate (WER). These metrics can be calculated using the equations \ref{eq:cer}, where $S$ represents the number of substitutions, $D$ represents the number of deletions, $I$ represents the number of insertions, and $C$ represents the total number of characters in the reference text. Similarly, WER can be calculated using equation \ref{eq:wer}, where $N$ represents the total number of correct words. A lower value indicates better performance.

\begin{equation}
\label{eq:cer}
    CER = \frac{S + D + I}{S + D + I + C}
\end{equation}

\begin{equation}
\label{eq:wer}
    WER = \frac{S_w + D_w + I_W}{N_w}
\end{equation}

\paragraph{Evaluation on OCR Models with GPT}Table \ref{tab:evaluation-word-level-text} presents the performance of various OCR models on the word-level correctness of the ICDAR2013 dataset (about more than 800 images), considering both the original ground truth and the fixed ground truth. Our analysis focuses on EasyOCR, a slightly inferior OCR model, as well as two superior OCR models, PaddleOCR and TrOCR. Utilizing gpt-3.5-turbo-0301 for post-OCR correction demonstrates a noticeable improvement in the OCR results, indicating that the language model can enhance and compensate for the deficiencies to some extent.\par
In Table \ref{tab:evaluation-line-level-text}, We demonstrate the post-OCR correction process based on ICDAR 2023 at line-level (we have selected about 50 images with hierarchical text). Similarly, the language model proves beneficial in rectifying errors by leveraging contextual information.

\paragraph{Evaluation on Glyph Embedding Models} Table \ref{tab:open-set} illustrates the performance of various models used for training glyph embedding. Unfreezing more than the two modified fc layers can lead to disturbances in the pre-trained ResNet model, resulting in decreased accuracy. Furthermore, in our specific case, the garbage class approach yields superior results compared to the adapted softmax approach.
\paragraph{Evaluation on Post-OCR Correction Models}
Table \ref{tab:post_ocr} demonstrates that the inclusion of glyph embedding leads to a significant enhancement in the WER and CER scores at the sentence level. However, the impact on the performance at the single-word level is minimal. Notably, the model exhibits proficient post-correction capabilities at the single-word level, as evidenced by the low CER and WER scores.
\section{Discussion}
\label{sec:discussion}
\paragraph{}During the initial phase of the study, we observed that post-OCR correction possesses the capacity to mitigate the limitations of OCR models to a certain degree. This becomes particularly evident when processing inputs comprising complete sentences. The application of post-OCR correction by GPT to both PaddleOCR and TrOCR outputs resulted in notable and comparable improvements. However, in scenarios involving inputs containing a single word, such as those generated by EasyOCR, the impact of post-correction on performance enhancement is relatively minimal.\par
In the subsequent step, the comparative performance of the post-OCR correction models developed by us is examined. Notably, these models achieve results comparable to those generated by GPT (Table \ref{tab:evaluation-line-level-text}), while employing a significantly lighter weight model. It is worth mentioning that the aforementioned models are capable of effectively post-correcting individual words. Additionally, the inclusion of the glyph embedding leads to a substantial improvement in overall performance for all inputs.\par
However, the scope of this study is limited to characters within the ranges of 0-9, a-z, and A-Z. As a result, punctuations have not been included in the training process or considered during the evaluation. Nevertheless, this issue can be addressed and improved by incorporating additional training data for special characters and punctuations into the glyph embedding framework.
\begin{table}
\centering
\begin{tabular}{lrr}
\hline
\textbf{Layer} & \textbf{in dim} & \textbf{out dim}\\
\hline
\hline
Resnet50 last fc & 2048 & 768\\
additional fc & 768 & 63\\
\hline
\end{tabular}
\caption{\label{tab:resnet_arch}
The dimensions of the last two layers of the modified ResNet50 architecture were adjusted for the pre-trained task.
}
\end{table}
\begin{table}
\centering
\begin{tabular}{lll}
\hline
\textbf{Model} & \textbf{Original GT} & \textbf{Fixed GT}\\
\hline
\hline
EasyOCR & 0.788 & 0.800 \\
PaddleOCR & 0.948 & 0.958 \\
TrOCR & \textbf{0.952} & \textbf{0.965} \\
\hline
EasyOCR+GPT* & - & 0.841 \\
PaddleOCR+GPT* & - & 0.967 \\
TrOCR+GPT* & - & \textbf{0.969} \\
\hline
\end{tabular}
\caption{\label{tab:evaluation-word-level-text}
The evaluation of word-level text was performed using various models, with accuracy calculated based on full matching. The term "Original GT" denotes the ground truth initially provided by the dataset, while recognizing that it may contain errors. Consequently, we manually rectified some of these errors, resulting in a revised ground truth referred to as "Fixed GT". "*GPT" specifically represents the GPT-3.5-turbo-0301 model.
}
\end{table}
\begin{table}
\centering
\begin{tabular}{lll}
\hline
\textbf{Model} & \textbf{WER} & \textbf{CER}\\
\hline
\hline
EasyOCR & 0.67 & 0.58 \\
PaddleOCR & 0.45 & \textbf{0.11} \\
TrOCR & \textbf{0.22} & 0.12 \\
\hline
EasyOCR+GPT* & 0.66 & 0.55 \\
PaddleOCR+GPT* & \textbf{0.11} & \textbf{0.08} \\
TrOCR+GPT* & \textbf{0.11} & 0.09 \\
\hline
\end{tabular}
\caption{\label{tab:evaluation-line-level-text}
The evaluation of line-level text was conducted using diverse models, with accuracy measured through the utilization of Word Error Rate (WER) and Character Error Rate (CER). In this context, "*GPT" specifically denotes the GPT-3.5-turbo-0301 model.
}
\end{table}
\begin{table}
\centering
\begin{tabular}{lr}
\hline
\textbf{Model} & \textbf{Acc (\%)}\\
\hline
\hline
RN18, garbage class, unfreeze L2 & 85.08 \\
RN50, garbage class, unfreeze L2 & \textbf{87.37} \\
RN50, garbage class, unfreeze L4 & 83.90 \\
RN50, adapted softmax, unfreeze L4 & 83.19 \\

\hline
\end{tabular}
\caption{\label{tab:open-set}
This table shows models employed for training glyph embedding and provides a comprehensive comparison of their performance.
(RN refers to ResNet; L refers to number of layers)}
\end{table}
\begin{table}
\centering
\begin{tabular}{llrr}
\hline
\textbf{OCR Output} & \textbf{Glyph} & \textbf{WER} & \textbf{CER}\\
\hline
\hline
EasyOCR & False & - & 0.0959 \\
EasyOCR & True & - & \textbf{0.0292}\\
\hline
PaddleOCR  & False & 0.2642 & 0.1771  \\
PaddleOCR  & True & \textbf{0.0630 }&\textbf{0.0283}   \\
\hline
TrOCR  & False & 0.0859 & 0.0478 \\
TrOCR  & True & \textbf{0.0067}&\textbf{0.0013}   \\
\hline
\end{tabular}
\caption{\label{tab:post_ocr}
The table provides a comprehensive comparison of post-OCR correction models that integrate glyph embedding with those that exclusively employ CharBERT embedding. The comparison encompasses various OCR model outputs.
}
\end{table}
\bibliography{anthology,custom}
\bibliographystyle{acl_natbib}

\end{document}